\documentclass[runningheads]{llncs}
\usepackage{hyperref}
\usepackage{amsmath,amssymb,amsfonts}
\usepackage{algorithmic}
\usepackage{graphicx}
\usepackage{xcolor}
\usepackage{nicefrac}
\usepackage{float}
\usepackage{booktabs}
\def\BibTeX{{\rm B\kern-.05em{\sc i\kern-.025em b}\kern-.08em
    T\kern-.1667em\lower.7ex\hbox{E}\kern-.125emX}}
\usepackage{tikz}
\usetikzlibrary{patterns}

\usepackage{subcaption}

\author{
Max Berrendorf\inst{1}\orcidID{0000-0001-9724-4009}
\and
Evgeniy Faerman\inst{1}\and
 Laurent Vermue\inst{2}\and
Volker Tresp\inst{1,3}}
\authorrunning{M. Berrendorf et al.}
\institute{
Ludwig-Maximilians-Universität München, Munich, Germany\\ \email{\{berrendorf,faerman\}@dbs.ifi.lmu.de}
\and
Technical University of Denmark, Kongens Lyngby, Denmark\\\email{lauve@dtu.dk}
\and
Siemens AG, Munich, Germany\\
\email{volker.tresp@siemens.com}
}

\title{
On the Ambiguity of Rank-Based Evaluation of Entity Alignment or Link Prediction Methods
}

\titlerunning{
On the Ambiguity of Rank-Based Evaluation of EA or LP Methods
}

\begin{document}
\maketitle
\begin{abstract}
In this work, we take a closer look at the evaluation of two families of methods for enriching information from knowledge graphs:
Link Prediction and Entity Alignment.
In the current experimental setting, multiple different scores are employed to assess different aspects of model performance.
We analyze the informativeness of these evaluation measures and identify several shortcomings.
In particular, we demonstrate that all existing scores can hardly be used to compare results across different datasets.
Moreover, we demonstrate that varying size of the test size automatically has impact on the performance of the same model based on commonly used metrics for the Entity Alignment task.
We show that this leads to various problems in the interpretation of results, which may support misleading conclusions. 
Therefore, we propose adjustments to the evaluation and demonstrate empirically how this supports a fair, comparable, and interpretable assessment of model performance.
Our code is available at \url{https://github.com/mberr/rank-based-evaluation}.
\end{abstract} \section{Introduction}
Information retrieval systems often require information organized in an easily accessible and interpretable structure.
Frequently, Knowledge Graphs (KGs) are used as an information source~\cite{Dietz2019}.
Consequently, the successful application of new information retrieval algorithms often depends on the completeness and quality of the information in KGs.
\emph{Link Prediction}~(LP)~\cite{nickel2015review} and \emph{Entity Alignment}~(EA)~\cite{berrendorf2019knowledge} are two disciplines with the goal to enrich information in KGs.
LP makes use of existing information in a single KG by materializing latent links.
The goal of EA is to align entities in different KGs, which facilitates the transfer of information between both or a fusion of multiple KGs to a single knowledgebase.
Both disciplines work by assigning scores to potential candidates:
LP methods compute scores for the facts in question at inference time and EA methods assign scores to candidate alignment pairs. 
Simple thresholding, or also more advanced assignment methods~\cite{kuhn1955hungarian} for EA, can make use of these scores to predict new links or alignments.

During the evaluation, both, LP and EA, evaluate how the "true" entity is \emph{ranked} relative to other candidate entities.
Given a \emph{rank} for each test instance, various metrics exist to obtain a single number quantifying the overall performance of an approach.
In this paper, we analyze the whole evaluation procedure and make the following contributions:
\begin{enumerate}
    \item While the rank of the true entity is an intuitive and simple concept, after reviewing numerous existing codebases, we found several competing methods to compute a rank. 
    \item We describe the intuition behind current aggregation scores and argue that they do not always provide a complete picture of the model performance. We show that this is an actual problem in the current evaluation setting, which sometimes may lead to wrong conclusions. 
    \item We propose a new (adapted) evaluation score overcoming the problems, and empirically demonstrate its usefulness for both tasks.
\end{enumerate}
The remainder of the paper is structured as follows:
In Section~\ref{sec:framework}, we discuss existing rank definitions, and aggregation metrics summarizing individual ranks.
In Section~\ref{sec:ours}, we point out the problems of current evaluation and introduce an adapted aggregation metric, which circumvents the shortcomings of existing aggregations.
Afterwards, in Section~\ref{sec:related_work}, we discuss related work.
Finally, in Section~\ref{sec:experiments}, we demonstrate empirically the shortcomings in the current evaluation protocol as well as the effects of our adaptations and conclude in Section~\ref{sec:conclusion}.

\section{Evaluation Framework}
\label{sec:framework}
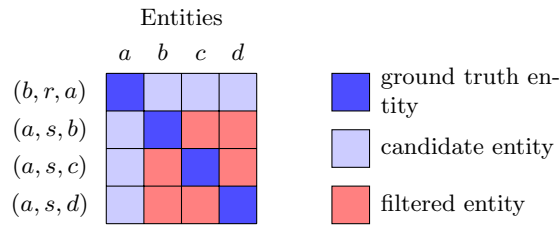
\begin{figure}
    \centering
    \begin{tikzpicture}[scale=.5, yscale=-1]
    \draw (2, -1.5) node {Entities};
    \foreach \x/\l in {0.5/a,1.5/b,2.5/c,3.5/d}{
        \draw (\x, -.5) node {\strut$\l$};
    }
    \foreach \y/\h/\r/\t in {0.5/b/r/a,1.5/a/s/b,2.5/a/s/c,3.5/a/s/d}{
        \draw (-1.5, \y) node {\strut$(\h, \r, \t)$};
    }
    \colorlet{t}{blue!70}
    \colorlet{y}{blue!20}
    \colorlet{n}{red!50}
    \foreach \x/\y/\c in {0/0/t,1/0/y,2/0/y,3/0/y,0/1/y,1/1/t,2/1/n,3/1/n,0/2/y,1/2/n,2/2/t,3/2/n,0/3/y,1/3/n,2/3/n,3/3/t}{
        \fill[color=\c] (\x, \y) rectangle +(1, 1);
    }
    \draw (0, 0) grid +(4, 4);
    
    \fill[draw=black,fill=t] (6, 0) rectangle +(1, 1);
    \draw (10, 0.5) node[text width=2.7cm, align=left] {ground truth entity};
    \fill[draw=black,fill=y] (6, 1.5) rectangle +(1, 1);
    \draw (10, 2) node[text width=2.7cm, align=left] {candidate entity};
    \fill[draw=black,fill=n] (6, 3) rectangle +(1, 1);
    \draw (10, 3.5) node[text width=2.7cm, align=left] {filtered entity};
    \end{tikzpicture}
    \caption{
    Visualization of candidate sets for the filtered evaluation setting for link prediction (right side / tail prediction) on a toy example with triples $\{(b, r, a), (a, s, b), (a, s, c), (a, s, d)\}$.
    Depending on the presence of other triples with shared head-relation pairs, the number of considered candidate entities varies, and consequently the maximum possible rank.
    In this example, there are three triples starting with $(a,s)$. 
    When, e.g., triple $(a,s,b)$ is evaluated $c$ and $d$ are ignored.
    Since only two entities remain, the rank cannot be larger than two.
}
    \label{fig:example_lp}
\end{figure}

\subsection{Rank for Link Prediction and Entity Alignment}
\paragraph{Link Prediction}
Let a single knowledge graph be represented as $\mathcal{G} = (\mathcal{E}, \mathcal{R}, \mathcal{T})$, where $\mathcal{E}$ is a set of entities, $\mathcal{R}$ is a set of relations, and $\mathcal{T} \subseteq \mathcal{E} \times \mathcal{R} \times \mathcal{E}$ is a set of triples.
For the task of LP a set of given triples is usually divided in $\mathcal{T}_{train} \subseteq \mathcal{T}$ and $\mathcal{T}_{test} = \mathcal{T} \setminus \mathcal{T}_{train}$, where $\mathcal{T}_{test}$ is used to assess the model performance.
A common evaluation protocol is to use every triple $(h, r, t) \in \mathcal{T}_{test}$, and perform left-side and right-side prediction.
For the right-side prediction, the score for every triple $\{(h, r, e) \mid e \in \mathcal{E}\}$ is computed and the entities $e$ are sorted in decreasing order by the predicted scores.
The \emph{rank} of the "true" entity $t$ is computed as the index in the resulting sorted list.
The left-side prediction follows analogously.
The final rank of the triple is computed as an average over both ranks, left-side and right-side.
To account for the possibility of multiple existing links for a given head-relation / relation-tail pair, the filtered evaluation setting was introduced~\cite{Bordes2013}:
When scoring tail entities for a triple $(h, r, t)$, all other entities $t \neq t' \in \mathcal{E}$ with triples $(h, r, t') \in (\mathcal{T}_{train} \cup \mathcal{T}_{test})$ are ignored, cf. Figure~\ref{fig:example_lp}.
Therefore, the performance does not decrease when other entities are scored higher than the currently considered one, as long as they are also true.
The filtered evaluation protocol is the quasi-standard for link prediction on knowledge graphs, and unfiltered scores are rarely reported.

\paragraph{Entity Alignment}
\begin{figure*}
    \centering
\begin{tikzpicture}[scale=.5, yscale=-1]
    
    \fill[color=red!20] (0, 1) rectangle +(2, 1);
    \fill[color=red!60] (0, 3) rectangle +(2, 2);
    
    \draw (1, 1.5) node {A};
    \draw (1, 4) node {C};
    
    \fill[color=blue!20] (3, 1) rectangle +(2, 1);
    \fill[color=blue!60] (3, 3) rectangle +(2, 2);
    
    \draw (4, 1.5) node {B};
    \draw (4, 4) node {D};
    
    \draw[color=red] (0, 0) rectangle +(2, 7);
    \draw[color=blue] (3, 0) rectangle +(2, 6);
    
    \draw[color=red] (1, -0.5) node {$\mathcal{E}_L$};
    \draw[color=blue] (4, -0.5) node {$\mathcal{E}_R$};
    
    \draw (-1, 1) rectangle +(7, 1);
    \draw (-1, 3) rectangle +(7, 2);
    
    \draw (7.2, 1.5) node {$\mathcal{A}_{train}$};
    \draw (7, 4) node {$\mathcal{A}_{test}$};
    
    \draw (-1.5, 1) edge [latex-latex] +(0, 1);
    \draw (-3, 1.5) node {$|\mathcal{A}_{train}|$};
    \draw (-1.5, 3) edge [latex-latex] +(0, 2);
    \draw (-3, 4) node {$|\mathcal{A}_{test}|$};
    
    \fill[color=red!20] (12.5, 0) rectangle +(1, 2);
    \fill[color=blue!20] (10, 2.5) rectangle +(2, 1);
    \fill[color=red!50!blue!20] (12.5, 2.5) rectangle +(1, 1);
    
    \draw (13, 1) node {A};
    \draw (11, 3) node {B};
    
    \fill[color=red!60] (15, 0) rectangle +(2, 2);
    \fill[color=blue!60] (17.5, 2.5) rectangle +(2, 2);
    \fill[color=red!50!blue!60] (15, 2.5) rectangle +(2, 2);
    
    \draw (16, 1) node {C};
    \draw (18.5, 3.5) node {D};
    
    \draw (13, 7) node[text width=6cm] (text) {Number of candidates for each entity differs for $\mathcal{A}_{train}$ vs. $\mathcal{A}_{test}$.};
    
    \draw (text.north) edge[-latex] (13, 3.5);
    \draw (text.north) edge[-latex] (15, 4.5);
    
    \end{tikzpicture}
    \caption{
    Visualization of the commonly used evaluation protocol for entity alignment.
    When evaluating performance for a given ground truth alignment $\mathcal{A}_{eval}$, only those entities are considered as candidates that occur in any pair of $\mathcal{A}_{eval}$.
    Thus, with varying size $|\mathcal{A}_{eval}|$, the number of candidates varies as well.
    }
    \label{fig:example_ea}
\end{figure*}
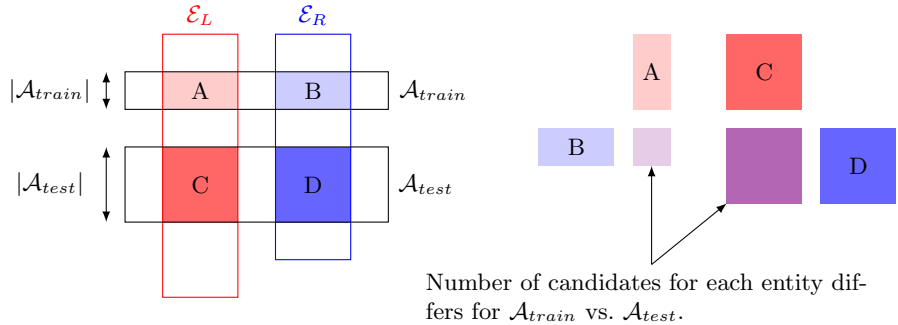
For this task, there are two knowledge graphs $\mathcal{G}_L = (\mathcal{E}_L, \mathcal{R}_L, \mathcal{T}_L)$ and $\mathcal{G}_R = (\mathcal{E}_R, \mathcal{R}_R, \mathcal{T}_R)$, and a set of aligned entities $\mathcal{A} \subseteq \mathcal{E}_L \times \mathcal{E}_R$.
Analogous to the previous evaluation setting, the set of alignments is divided into $\mathcal{A}_{train} \subseteq \mathcal{A}$ and $\mathcal{A}_{test} = \mathcal{A} \setminus \mathcal{A}_{train}$. The common evaluation scheme \cite{sun2017cross,chen2017multigraph,sun2018bootstrapping,wang2018cross,pei2019semi,xu2019crosslingual,cao2019multi,DBLP:conf/icml/GuoSH19,trisedya2019entity,zhu2019neighborhood,zhang2019multi,sun2019knowledge} now computes scores for every candidate pair $\{(a_L, e_R) \mid e_R \in \mathcal{E}_R, \exists a_L' \in \mathcal{E}_L: (a_L', e_R) \in \mathcal{A}_{test}\}$, and determines the rank of the "true" score of $(a_L, a_R)$.
The right-side prediction is defined correspondingly.
Notice, that only those entities are considered for which there exists an aligned entity from the other graph in the \emph{test} part of the alignment.

\subsection{Variants of Rank}
\label{subsec:rank}
To compute a rank in LP and EA evaluation, we are given a list of scores $\mathcal{S} = [\beta_1, \ldots, \beta_C]$ for each test instance  with $|\mathcal{C}| = C$, where $\mathcal{C}$ is a set of candidates.
We denote the score of the "true" entity as $\alpha$, and its position in the decreasingly sorted list as $r=\textit{rank}(S, \alpha)$.
As also noted by \cite{sun2019re}, there exist different variants to compute the rank, essentially differing in their way of how they handle equal scores.
In the following, we provide a taxonomy and refer to papers that use the corresponding definition\footnote{Usually, the exact way how the scores are computed is not detailed in the paper itself but found in the code published alongside. 
In the appendix you can find links to the source code of evaluation for all cited works.
}.

\paragraph{Optimistic rank}
The optimistic rank assumes that the object in question is ranked first among those with equal score, i.e.,
\begin{equation}
    rank^+(\mathcal{S}, \alpha) := \left|\{\beta \in \mathcal{S} \mid \beta > \alpha\}\right| + 1.
\end{equation}
This definition is for instance used in the implementations of~\cite{Han2018,Kazemi2018,Lerer2019,Liu2017,Trouillon2016,chen2017multigraph,pei2019semi,Yu2019}.

\paragraph{Pessimistic rank}
The pessimistic rank assumes that the object in question is ranked last among those with equal score, i.e.,
\begin{equation}
rank^-(\mathcal{S}, \alpha) := \left|\{\beta \in \mathcal{S} \mid \beta \geq \alpha\}\right|.    
\end{equation}
This definition is for instance used in the implementations of~\cite{Lacroix2018,Schlichtkrull2018}.

\paragraph{Non-Deterministic rank}
The non-deterministic rank applies any sorting algorithm to the list and uses the index in this list as rank.
Thus, the value depends on the inner workings of the applied sort algorithm, e.g. its sort stability, as well as the initial ordering of the list, i.e. the order in which the ranks where computed.
Note, that it is different from the \emph{random} rank proposed in~\cite{sun2019re} since it is non-deterministic regarding different implementations of sorting algorithms.
This definition is for instance used in the implementations of~\cite{Balazevic2018,Balazevic2019,Balazevic2019b,Dasgupta2018,Dettmers2018,Nickel2016,sun2018bootstrapping,wang2018cross,sun2019knowledge,sun2018bootstrapping,sun2017cross,xu2019crosslingual,cao2019multi,trisedya2019entity,zhang2019multi}. 

\paragraph{Realistic rank}
The realistic rank is given as the mean of optimistic and pessimistic rank, also equal to the average over all valid ranks, i.e. the positions the entity can be placed within the list without violating the sort criterion.
\begin{equation}
rank(\mathcal{S}, \alpha) := \frac{1}{2}(rank^+(\mathcal{S}, \alpha) + rank^-(\mathcal{S}, \alpha))
\end{equation}
This definition is for instance used in the implementations of~\cite{Ding18,Jin2019,DBLP:conf/icml/GuoSH19}.

\subsection{Overall metrics}
\label{subsec:overall_metrics}
Given the set of individual rank scores
$\mathcal{I}$, the following scores are commonly used as aggregation.
\subsubsection{Hits @ k}
The \emph{Hits @ k} (H@k) score describes the fraction of hits, or fraction of instances, for which the "true" entity appears under the first $k$ entities in the sorted list:
\begin{equation}
H@k := \frac{|\{r \in \mathcal{I} \mid r \leq k\}|}{|\mathcal{I}|}.    
\end{equation}
In the context of information retrieval, this metric is also known as Precision@k.
One of the advantages of this metric is that it is easily interpretable.
Since for many applications only the first outputs are taken into account, it can help to directly assess the method's applicability to the use-case.
However, this metric does not distinguish the cases, where the rank is larger than $k$.
Thus, the ranks $k+1$ and $k+d$, where $d \gg 1$, have the same effect on the final score.
Therefore, it is less suitable for the comparison of different models.

\subsubsection{Mean Rank}
The \emph{mean rank} (MR) computes the mean over all individual ranks:
\begin{equation}
\label{eq:mr}
MR := \frac{1}{|\mathcal{I}|} \sum \limits_{r \in \mathcal{I}} r.
\end{equation}
The advantage of the MR score is that it is sensitive to any model performance changes.
If the rank on the same evaluation set becomes better on average, the improvement is always reflected by the MR score.
While the MR is still interpretable, it is necessary to keep the size of the candidate set in mind to assess the model performance and interpret its value:
A MR of 10 might indicate strong performance for a candidate set size of 1,000,000, but for a candidate set of only 20 candidates it equal to the expected performance of a model with random scorings.

\subsubsection{MRR}
The \emph{mean reciprocal rank} is still often reported along with other scores.
It is defined as
\begin{equation}
    MRR := \frac{1}{|\mathcal{I}|} \sum_{r \in \mathcal{I}} \frac{1}{r}.
\end{equation}
While the MRR is less sensitive to outliers and has the property to be bounded in the range $(0, 1]$, it was shown that this metric has serious flaws and therefore should not be relied upon~\cite{fuhr2018some}.
However, especially in LP codebases, the MRR is often used for early stopping.
Presumably, the main reason for that is the behavior of the reciprocal function:
While the Hits@k score ignores change among high rank values completely, MR values changes uniformly among the full value range.
The MRR score, in contrast, is more affected by changes of low rank values than high ones, but it does not completely disregard them.
Therefore, it can be considered as soft a version of Hits@k.

\section{Our Evaluation Approach}
\label{sec:ours}
In the following, we briefly discuss our choice for the rank definition.
Subsequently, we point out problems in the current evaluation scheme and describe our new overall score.

\subsection{Optimal Rank}
We identified the following desiderata for the selection of optimal rank definition:
\begin{itemize}
    \item There are models which assign the same score to significant portions of positive and negative triples \cite{sun2019re}.
    The \emph{optimistic rank} is sensitive to these degenerate cases and gives such models an unfair advantage.
    \item Adequate assessment of model performance:
    Since the \emph{pessimistic} rank gives too conservative assessments it is not suitable, underestimating the model's usefulness.
    \item Reproducibility across different experiment runs and environments: The \emph{non-deterministic} or \emph{random} \cite{sun2019re} ranks are not always reproducible. 
\end{itemize}
Therefore, we propose to use the \emph{realistic} rank.
While being deterministic, the realistic rank offers a trade-off between the pessimistic and optimistic rank.
In the degenerate case, when $\beta_{i}=c \in \mathbb{R}$ for all $\beta_{i} \in \mathcal{S}$, the \emph{realistic} rank is equal to $\nicefrac{|\mathcal{S}|}{2}$, i.e. very close to the expected rank in the case, when all scores are drawn at random.
If $\alpha, \beta_1, \ldots, \beta_C$ are i.i.d and drawn at random, and therefore the element can appear at any position with the same probability, the expected rank is also the middle of the sorted array:
\begin{equation}
\label{eq:exp_rank}
\mathbb{E}[rank(\mathcal{S}, \alpha)]
= \frac{1}{n} \sum \limits_{i=1}^{|\mathcal{S}|} i
= \frac{1}{2}(|\mathcal{S}|+1)    
\end{equation}

\subsection{Adjusted Mean Rank}
While the \emph{H@k} score enables assessments of the model's suitability for a use-case, the \emph{MR} allows a more fine-grained comparison between different models.  Both metrics are necessary to get the entire picture of the model performance, e.g. the evaluation in~\cite{wang2019knowledge} demonstrates, that an excellent \emph{H@k} score does not necessarily coincide with a good \emph{MR}.
However, since the \emph{MR} score denotes the absolute position, it is not easily interpretable. Therefore, the comparison between experiments with different sizes of candidate sets is not easily possible with implications for the evaluation of both tasks.
\paragraph{Link Prediction}
The results on datasets with a different number of entities are not directly comparable.
However, comparability of performance on different datasets is important, for example, to assess the task complexity, choose benchmarks, or investigate model generalization.
For instance, surprisingly good test scores can be an indication for test leakage, see e.g.~\cite{toutanova2015observed}.
Intuitively, the number of candidates is an important factor directly affecting the task complexity, while it is not the only factor. 

\paragraph{Entity Alignment}
While the comparison of the performance on different datasets is also difficult for EA, there is the additional problem that only those entities are considered as candidates, which occur in at least one \emph{test} alignment. 
Therefore, the number of candidates depends on the size of the evaluation alignment set, cf. Figure~\ref{fig:example_ea}. 
Thus, results on the same dataset are not comparable for different train/test splits or between train and test sets.
This can lead to various misinterpretations of results.
For instance, in~\cite{wang2018cross,mao2020mraea}, the authors show an experiment where they increase the training size step-wise and evaluate the model on the rest of the data.
Based on the score improvement, they conclude that the model benefits from additional training data.
While this claim can still be true, we argue that another evaluation is necessary to support it.
As we demonstrate in Section~\ref{sec:experiments}, both MR and Hits@k scores improve \emph{automatically} as the test set becomes smaller, even if the model stays exactly the same.
The necessary condition for such an evaluation is either independence on candidate set size or the same candidate set for all experiments.
One possible solution would be to use all entities in the KG as candidates analogous to LP.
However, this still would leave us with the unresolved problem of performance comparison across datasets.
Therefore, we propose an adjustment to the \emph{MR} score that assesses the model performance independently of the candidate set size.

\paragraph{Adjusted Mean Rank Index}
Since we are interested in evaluating model performance, we start with the mean rank as our starting point.
Inspired by the Adjusted Rand Index~(ARI)~\cite{rand1971objective}, we aim to adjust it for chance.
Therefore, we compute the \emph{expected mean rank} following the assumption that the individual ranks are independent:
\begin{eqnarray*}
\mathbb{E}\left[MR\right]
&\overset{\eqref{eq:mr}}{=}& \mathbb{E}\left[\frac{1}{n} \sum \limits_{i=1}^{n} rank(\mathcal{S}_i, \alpha)\right]
= \frac{1}{n} \sum \limits_{i=1}^{n} \mathbb{E}\left[rank(\mathcal{S}_i, \alpha)\right]\\
&\overset{\eqref{eq:exp_rank}}{=}& \frac{1}{n} \sum \limits_{i=1}^{n} \frac{|\mathcal{S}_i|+1}{2}
= \frac{1}{2n} \sum \limits_{i=1}^{n} (|\mathcal{S}_i|+1)
\end{eqnarray*}
Now, we define the \emph{adjusted mean rank} as the MR divided by its expected value:
\begin{eqnarray*}
AMR 
= \frac{MR}{\mathbb{E}\left[MR\right]}
= \frac{2 \sum_{i=1}^{n} r_{i}}{\sum_{i=1}^{n} (|\mathcal{S}_i|+1)}
\end{eqnarray*}
Finally, to obtain a measure where 1 corresponds to optimal performance, we transform the adjusted mean rank to \emph{adjusted mean rank index} (AMRI) as follows:
\begin{equation}
AMRI 
= 1 - \frac{MR - 1}{\mathbb{E}\left[MR - 1\right]}
= 1 - \frac{2 \sum_{i=1}^{n} (r_{i} - 1)}{\sum_{i=1}^{n} (|\mathcal{S}_i|)}
\end{equation}
Since $r_{i} - 1 \leq |\mathcal{S}| - 1$ the AMRI has a bounded value range of $[-1, 1]$.
A value of 1 corresponds to optimal performance where each individual rank is 1.
A value of 0 indicates model performance similar to a model assigning random scores, or equal score to every candidate. The value is negative if the model performs worse than the constant-score model.

\section{Related Work}
\label{sec:related_work}
A special property of the \emph{ranking} evaluation is that the candidate scores for each test instance are only required to be comparable within a single candidate set.
The scores of candidates for another test instance may have a different value range, but since they are not compared with the candidates of other test instances, this does not affect the results.
Therefore, ranking evaluation is appropriate for a setting where a human can evaluate model proposals.
If, on the other hand, the decision has to be made automatically, e.g. using a fixed threshold, the \emph{classification} setting is more appropriate.
In the following, we review related approaches and demonstrate that classification and ranking evaluations are used interchangeably.

\subsection{Triple Classification}
In the LP task, we are given a pair of a head/tail entity $e \in \mathcal{E}$ and relation $r \in \mathcal{R}$, and rank a set of possible tail/head entities $e' \in \mathcal{E}$ according to the plausibility of the triple $(e, r, e')$ / $(e', r, e)$.
In contrast, for \emph{triple classification}, we aim at classifying whether a triple is true or false irrespective of the plausibility of other triples~\cite{DBLP:conf/nips/SocherCMN13,Wang2014,DBLP:conf/aaai/LinLSLZ15,DBLP:conf/ijcai/XieLS16,DBLP:conf/emnlp/GuoWWWG16}.
Consequently, a global threshold for the score of triples is required for the classification decision.
If the threshold is chosen manually, classification metrics such as accuracy or $F_1$-measure can be used.
Otherwise, the area under the precision-recall curve (PR-AUC), or receiver-operator curve (ROC-AUC) are used to summarize the performance over all possible decision thresholds.
Link prediction and triple classification are sometimes evaluated alongside to demonstrate the effectiveness of novel knowledge graph embedding models across different tasks~\cite{nickel2011three,krompass2013non,DBLP:conf/nips/SocherCMN13}.

\subsection{Ontology Matching}
\emph{Ontology matching} or \emph{instance matching} is closely related to EA.
Here we seek correspondences between instances of different ontologies based on different data properties of the instances.
In contrast to EA, the vast majority of ontology matching approaches are unsupervised, i.e. there are no training alignments, but the instance features that are used for matching~\cite{DBLP:conf/semweb/Jimenez-RuizG11,DBLP:conf/cikm/NiuRWY12,DBLP:journals/jcst/ShaoHLWCX16,belhadi2020data}.
The similarity is often fixed, e.g. to TF-IDF, and the methods optimize the matching process by pruning the candidate match space and selecting subsets of properties used for matching.
Ontology matching approaches are evaluated in a classification setting with precision/recall/$F_1$-measure as evaluation score~\cite{DBLP:conf/semweb/AlgergawyFFFHHJ19}.

\subsection{Information Retrieval}
The evaluation of LP and EA share some similarities with the evaluation of \emph{Information Retrieval} (IR) techniques. Given a query, the IR system returns a list of documents which should be of relevance to the query.
Different approaches are thus compared based on the "true" relevance of retrieved documents.
Earlier IR approaches were evaluated in the classification setting and \emph{precision} and \emph{recall} were reported for fixed candidate set for each query~\cite{blair1985evaluation,saracevic1988study}.
Later on, the evaluation was extended to summarize the evaluation for candidate sets of different sizes.
Therefore, the precision was evaluated multiple times after retrieving relevant documents and averaged~\cite{kishida2005property} or simply reported for multiple recall levels~\cite{zhang2009eleven}.
More recent evaluation protocols employ ranking approaches ~\cite{radev2002evaluating,sanderson2010christopher}.
Besides the ranking scores already discussed in the previous sections, gain-based approaches and especially the \emph{Normalized Discounted Cumulative Gain (NDCG)}~\cite{vakkari2000changes} are popular metrics.
These approaches associate a scalar usefulness, gain or importance level to each document.
The gain is then discounted by the position in the result list it appears at:
The later the relevant document appears the more its gain is discounted.
Since all true candidates are equally important in EA and LP tasks, we do not see any advantages by adapting gain based metric.
The problem of partial ordering in the result set was also actively investigated in the IR community for both classification and ranking evaluation~\cite{cooper1968expected,raghavan1989critical,mcsherry2008computing}.
While some competitions still use the \emph{non-deterministic rank}~\cite{voorhees2006trec}, McSherry et al.~\cite{mcsherry2008computing} demonstrate how the average over all possible orderings can be computed analytically for classification and ranking evaluation scores.

\section{Experiments}
\label{sec:experiments}
In this section, we evaluate empirically to what extent the results of the current evaluation protocol can be misinterpreted, and whether the adapted score helps to mitigate the problem.
First, we focus on the EA evaluation and train the same model with different amounts of training data.
Each individual model is then evaluated on test sets of different sizes. 
We use GCN-Align~\cite{wang2018cross} as model and the \texttt{zh-en} subset of the \texttt{DBP15k(JAPE)} dataset with the best hyperparameters from~\cite{berrendorf2019knowledge}.
\begin{figure}
\centering
\includegraphics[width=\linewidth,height=.8\textheight,keepaspectratio]{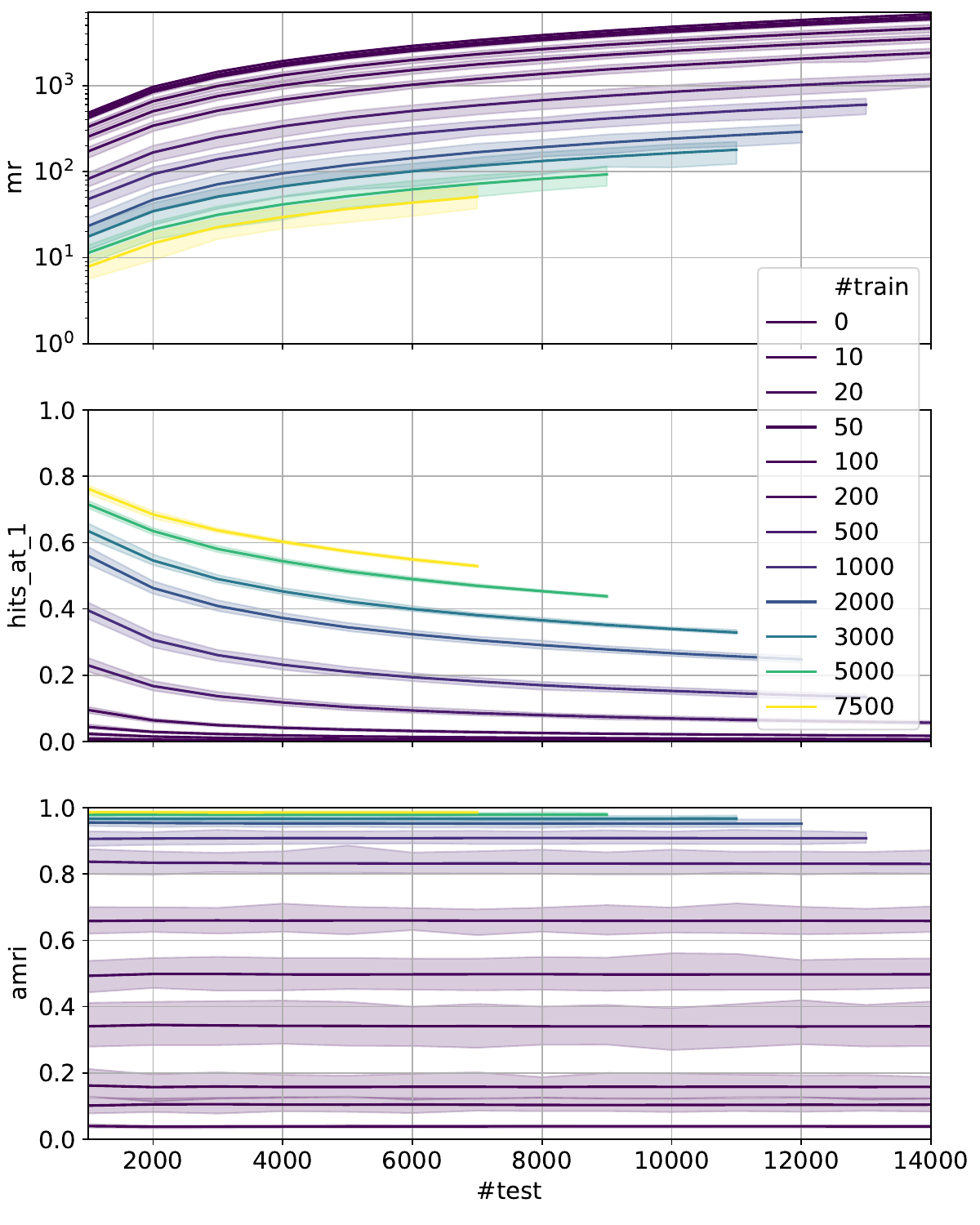}
  \caption{
  Results on \texttt{DBP15k(JAPE)}, \texttt{zh-en}.
  We train multiple models with different amounts of training data, indicated by the color of the line.
  Each of these models is then evaluated on a different number of test alignments (x-axis).
  Each row shows the evaluation results for one rank aggregation score on a logarithmic y-axis.
  The shaded area shows the variation across five different random train-test splits.
  Only AMRI shows consistent results for the same model with different number of test alignments.
  }
  \label{fig:three_metrics_comparison}
\end{figure}
\begin{figure}
    \centering
    \includegraphics[width=\linewidth]{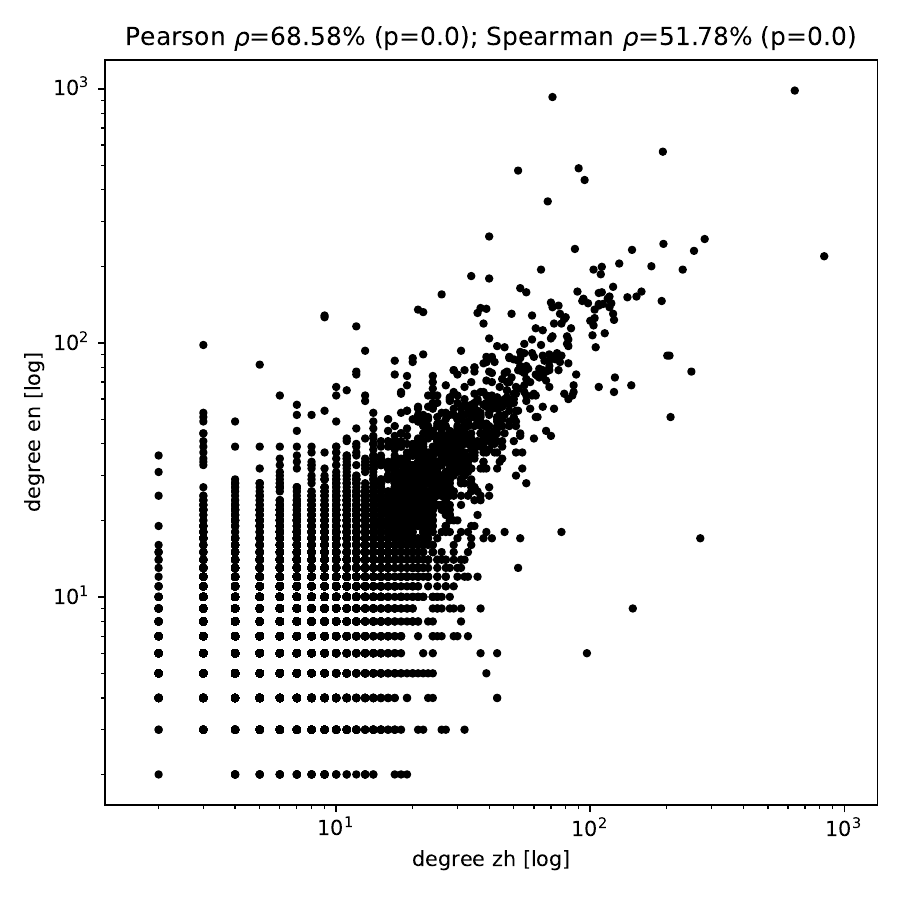}
    \caption{
    Node degrees and matching between graphs: 
    Each dot represents an aligned entity with the x/y coordinate corresponding to the degree in both graphs.
We observe that nodes tend to match to nodes of similar degree.
    }
    \label{fig:degree_prior}
\end{figure}
The results are presented in Figure~\ref{fig:three_metrics_comparison}.
These results confirm our assumptions about the behavior of the overall scores.
We observe that the MR increases almost linearly with increasing test size, when evaluated on exactly the same model (Note the log scale on Y-axis).
We also observe a similar effect for the Hits@1 score.
In contrast, the adjusted mean rank index (\emph{AMRI}) is almost completely insensitive to the size of the test set.
Now, we can also claim that GCN-Align takes advantage of more training alignments, since the models for larger number of training alignments have higher AMRI scores, although the gain dimishes with increasing number of training alignments.
However, we can also observe that additional training data is especially helpful if training set is small.
We also see that the results become more interpretable, e.g. we see that the models trained with 0 alignments, i.e. \emph{without any training}, already perform better than random.We hypothesize that this is due to the degree prior introduced by the message passing steps:
In Figure~\ref{fig:degree_prior}, we show a scatter plot of the node degrees of shared nodes in each graph.
We can observe that matching nodes tend to have a similar degree.
While the edge weights in GCN-Align are normalized such that the weight of incoming messages always sum up to one, the nodes with higher degree receive more (small) random vectors: 
Hence, the sum of them tends more towards the mean, zero, and the node representations of high-degree nodes tend to be more similar to each other than to low-degree nodes.
Using the Spearman rank correlation test, we obtain a correlation between the norm of the node representations and the node degree of $\rho\approx-0.041$ ($p\approx9.22 \cdot10^{-17})$ indicating a very small, but significant correlation.
The norm of the node embeddings without enrichment by message passing does not exhibit this correlation ($\rho\approx0.001$, $p\approx0.834$), and, as expected, the AMRI for evaluating those un-refined embeddings is close to zero ($-0.0025\% \pm0.8068 \%$).
We leave a more thorough verification of this process for future work.

\begin{table}
\centering
\caption{Link prediction results}
\label{tab:link_prediction}
\begin{tabular}{lrrrr}
\toprule
dataset & \multicolumn{2}{l}{WN18RR} & \multicolumn{2}{l}{FB15k-237} \\
metric &     MR &  AMRI (\%) &        MR &  AMRI (\%) \\
\midrule
DistMult~\cite{Yang2015}       &  7,000 & 65.8 &       500 & 93.0 \\
ConvE~\cite{Dettmers2018}      &  4,412 & 78.4 &       241 & 96.6 \\
TransE~\cite{Bordes2013}       &  2,289 & 88.8 &       317 & 95.6 \\
TransH~\cite{Wang2014}         &  2,126 & 89.6 &       219 & 97.0 \\
R-GCN~\cite{Schlichtkrull2018} &  6,254 & 69.4 &       540 & 92.5 \\
MuRP~\cite{Balazevic2019b}     &  2,448 & 88.0 &       167 & 97.7 \\
\bottomrule
\end{tabular}
\end{table}
In Table~\ref{tab:link_prediction}, we additionally compare the results of the \emph{LP} evaluation in the filtered setting on two datasets, for which we used evaluation results from~\cite{wang2019knowledge}, and also computed results for MuRP~\cite{Balazevic2019b} using their published code\footnote{https://github.com/ibalazevic/multirelational-poincare}.
Given the AMRI score we can clearly conclude that all methods perform better than random.
We also can compare the performance of the methods across datasets and observe consistently worse performance on the \emph{WN18RR} dataset.
This difference is not only due to the larger number of entities in WN18RR ($\approx45k$) compared to FB15k-237 ($\approx15k$), but has to be caused by a different mechanism, e.g. the higher sparsity of WN18RR, or the richer relational patterns in FB15k-237.
We leave the detailed analysis of dataset complexity for the future work.

\section{Conclusion}
\label{sec:conclusion}
In this work, we address problems in the evaluation of LP and EA models for knowledge graphs.
We thoroughly analyzed the current evaluation framework and identified several vulnerabilities.
We demonstrated their causes and effects and showed how the problems can be mitigated by a simple adjustment for chance.
Our empirical evaluation confirms our findings on both tasks. \bibliographystyle{splncs04}
\bibliography{references}
\clearpage
\section*{Appendix}
\small
\subsection{Optimistic} First among equal score.
\begin{itemize}
    \item ComplEx \cite{Trouillon2016} \\
    \url{https://github.com/ttrouill/complex/blob/dc4eb93408d9a5288c986695b58488ac80b1cc17/efe/evaluation.py#L295}
    \item OpenKE \cite{Han2018} \\
    \url{https://github.com/thunlp/OpenKE/blob/adeed2c0d2bef939807ed4f69c1ea4db35fd149b/base/Test.h#L64}
    \item PyTorch-BigGraph \cite{Lerer2019} \\
    \url{https://github.com/facebookresearch/PyTorch-BigGraph/blob/929dae01fe1a12d583733326e98139c33fc24690/torchbiggraph/eval.py#L59}
    \item SimplE \cite{Kazemi2018} \\
    \url{https://github.com/Mehran-k/SimplE/blob/29108230b63920afa38067b1aff8b8d53d07ed01/reader.py#L148}
    \item ANALOGY \cite{Liu2017} \\
    \url{https://github.com/quark0/ANALOGY/blob/fa9b6d397ec18c7282f20392afd757536c3d02a4/main.cpp#L425}
    \item MTransE \cite{chen2017multigraph} \\
    \url{https://github.com/muhaochen/MTransE/blob/96d17a266c97831a1c7e854320b222605fb9547c/src/TransE/TransE.py#L302-L309}
    \item SEA \cite{pei2019semi} \\
    \url{https://github.com/scpei/SEA/blob/4a5bc9535407d0a6ceb240f59275f6adb7fb808e/src/tester_SEA2.py#L330}
    \item Pykg2vec \cite{Yu2019}\\
    \url{https://github.com/Sujit-O/pykg2vec/blob/9ee1bb3543702af09f847fd98fdb983271fc3267/pykg2vec/utils/evaluator.py#L70-L96}
\end{itemize}

\subsection{Pessimistic} Last among equal score.
\begin{itemize}
    \item knowledge-graph-embeddings\\
    \url{https://github.com/mana-ysh/knowledge-graph-embeddings/blob/b24e41a4c62758c6645d580d55d72543a9ec3a82/src/processors/evaluator.py#L166}
    \item ComplEx-N3 \cite{Lacroix2018}\\
    \url{https://github.com/facebookresearch/kbc/blob/248e93136448661cac54020054b929997394f3d2/kbc/models.py#L68}
    \item R-GCN \cite{Schlichtkrull2018} \\
    \url{https://github.com/MichSchli/RelationPrediction/blob/c77b094fe5c17685ed138dae9ae49b304e0d8d89/code/common/evaluation.py#L151}
    \item graphvite\\
    \url{https://github.com/DeepGraphLearning/graphvite/blob/5cf3a5adcdc5252c04d6787d6784d4861154bc29/python/graphvite/application/application.py#L1061}
\end{itemize}

\subsection{Realistic}
\begin{itemize}
    \item ComplEx-NNE-AER \cite{Ding18}\\ \url{https://github.com/iieir-km/ComplEx-NNE_AER/blob/47fb01b3fa876a3d5154fa9b4e3c2b9fc2f29f08/ComplEx-NNE-AER/src/complex/Evaluation.java#L81}
    \item RENet \cite{Jin2019} \\
    \url{https://github.com/INK-USC/RE-Net/blob/36b2b832b467eb8f370a323deefdb94bba664781/model.py#L294}
    \item RSN \cite{DBLP:conf/icml/GuoSH19} \\
    \url{https://github.com/nju-websoft/RSN/blob/8025dd0ab6e291cdad114632f96fe49ab594f2e1/RSN4EA.ipynb}
\end{itemize}

\subsection{Non-Determistic}
\begin{itemize}
    \item ConvE \cite{Dettmers2018}\\
    \url{https://github.com/TimDettmers/ConvE/blob/5feb358eb7dbd1f534978cdc4c20ee0bf919148a/evaluation.py#L67}
    \item HolE \cite{Nickel2016}\\
    \url{https://github.com/mnick/holographic-embeddings/blob/c2db6e1554e671ab8e6acace78ec1fd91d6a4b90/kg/base.py#L198}
    \item Bootstrapping Entity Alignment \cite{sun2018bootstrapping}\\
    \url{https://github.com/nju-websoft/BootEA/blob/9861465e8114f666efc842464a269d330e3fb6c1/code/test_funcs.py#L50}
    \item HypER \cite{Balazevic2018} \\
    \url{https://github.com/ibalazevic/HypER/blob/1688b732047ad96cfae120748debb06a2fb7e209/HypER/hyper.py#L83}
    \item TuckER \cite{Balazevic2019} \\
    \url{https://github.com/ibalazevic/TuckER/blob/23aff592115ee91e099ce490cdbea28ef026fc54/main.py#L78}
    \item knowledge\_representation\_pytorch \\
    \url{https://github.com/jimmywangheng/knowledge_representation_pytorch/blob/26446f26ac16d57927da29c123718f244085dec4/evaluation.py#L183}
    \item DGL - R-GCN\\
    \url{https://github.com/dmlc/dgl/blob/2b8eb5be6fe11514ff504cf1cefc41f7a8e90a85/examples/pytorch/rgcn/utils.py#L176-L180}
    \item HyTE \cite{Dasgupta2018} \\
    \url{https://github.com/malllabiisc/HyTE/blob/96fc3498d3f3fbc7acd61d81a704537b239ac94d/result_eval.py#L39-L52}
    \item GCN-Align \cite{wang2018cross} \\ \url{https://github.com/1049451037/GCN-Align/blob/873d0c390bda67892300460b5904fc649e754cac/metrics.py#L61-L62}
    \item AliNet \cite{sun2019knowledge} \\ \url{https://github.com/nju-websoft/AliNet/blob/9ea135a4dd39dd471ee29a21402a541412a0197f/code/align/test.py#L110}
    \item BootEA \cite{sun2018bootstrapping} \\ 
    \url{https://github.com/nju-websoft/BootEA/blob/9861465e8114f666efc842464a269d330e3fb6c1/code/test_funcs.py#L22}
    \item JAPE \cite{sun2017cross} \\
    \url{https://github.com/nju-websoft/JAPE/blob/b4b3617a7c61df5f7093921553dd3b0a7497506d/code/embed_func.py#L180}
    \item \cite{xu2019crosslingual} \\
    \url{https://github.com/syxu828/Crosslingula-KG-Matching/blob/56710f8131ae072f00de97eb737315e4ac9510f2/run_model.py#L164-L178}
    \item MuGNN \cite{cao2019multi} \\
    \url{https://github.com/thunlp/MuGNN/blob/10113ea6a4f155a9d02c6129485588f1bfd015a7/utils/functions.py#L28}
    \item \cite{trisedya2019entity} \\
    \url{https://bitbucket.org/bayudt/kba/src/35c67d56a8f0e2bcc05e8d56fb98c4374e3ff542/KBA.ipynb#lines-496}
    \item MultiKE \cite{zhang2019multi} \\
    \url{https://github.com/nju-websoft/MultiKE/blob/a210a0c638ef4d91562bf098acb7153028dd74fc/code/base/alignment.py#L155}
\end{itemize}
 
\end{document}